\documentclass[a4paper, 10pt, conference, final]{ieeeconf}      
\usepackage{import}
\usepackage{subcaption}
\usepackage{threeparttable}
\usepackage{amsmath}
\usepackage{amssymb}  
\usepackage{bm}
\usepackage{makecell}
\usepackage{multirow}
\usepackage[table]{xcolor}
\usepackage{balance}
\usepackage{csquotes}

\usepackage{todonotes}
\IEEEoverridecommandlockouts                              

\title{\LARGE \bf

Safe and Psychologically Pleasant Traffic Signal Control \\with Reinforcement Learning using Action Masking
}

\author{Arthur Müller$^{1}$ and Matthia Sabatelli$^{2}$
\thanks{*This work is part of the KI4LSA project (19F2109A) and was supported by the German Federal Ministry for Digital and Transport (BMDV).}
\thanks{$^{1}$A. Müller is with the Fraunhofer Center for Machine Learning and Fraunhofer IOSB-INA, 32657 Lemgo, Germany:
        {\tt\footnotesize arthur.mueller@iosb-ina.fraunhofer.de}}
\thanks{$^{2}$M. Sabatelli is with the Department of Artificial Intelligence and Cognitive Engineering, University of Groningen, 9712 CP Groningen, The Netherlands:
        {\tt\footnotesize m.sabatelli@rug.nl}}
}

\begin{document}

\maketitle
\thispagestyle{empty}
\pagestyle{empty}

\begin{abstract} 
Reinforcement learning (RL) for traffic signal control (TSC) has shown better performance in simulation for controlling the traffic flow of intersections than conventional approaches. However, due to several challenges, no RL-based TSC has been deployed in the field yet. 
One major challenge for real-world deployment is to ensure that all safety requirements are met at all times during operation. 
We present an approach to ensure safety in a real-world intersection by using an action space that is safe by design. The action space encompasses traffic phases, which represent the combination of non-conflicting signal colors of the intersection. Additionally, an action masking mechanism makes sure that only appropriate phase transitions are carried out.
Another challenge for real-world deployment is to ensure a control behavior that avoids stress for road users. We demonstrate how to achieve this by incorporating domain knowledge through extending the action masking mechanism.  
We test and verify our approach in a realistic simulation scenario. 
By ensuring safety and psychologically pleasant control behavior, our approach drives development towards real-world deployment of RL for TSC.  
\end{abstract}

\section{Introduction}
Inefficient use of existing street infrastructure, which leads to congestion, is a major problem for humans and the environment. The cost of congestion, for example, was estimated at \$179 billion in urban areas of the United States in 2017 \cite{Schrank2019}. One reason for this inefficiency is the poor quality of intersection traffic signal controllers (TSC). Most of them use conventional control approaches, which encompass a set of rules to manage traffic flow depending on the current traffic volume. However, these rule-based approaches control traffic suboptimally in many cases.


One promising approach to overcome the limitations of conventional TSC is reinforcement learning (RL). In RL an agent
learns to act optimally in an environment to achieve an optimization goal. RL-based TSC has received a lot of research attention. Over the last years, a number of studies have demonstrated the superiority of RL over conventional approaches in simulation environments \cite{Haydari2020, Yau2017}. But to this date, there is no real-world deployment of an RL-based TSC, since there are several challenges to overcome when moving an RL-based controller from simulation to reality. One major challenge is to ensure that all safety requirements are met at all times during operation. 
Another challenge is to consider the psychological impact of the control behavior on the drivers so that driver stress is avoided


In this work, we address both challenges. First, we demonstrate how safety-related requirements of a TSC can be represented by a temporal, directed graph. Then, we incorporate this graph into an RL algorithm through action masking, which is a technique from the field of action space shaping \cite{Kanervisto2020}. Additionally, we define the action space to encompass discrete traffic phases representing legal combinations of signal colors, so that the action space is safe-by-design. Together with the incorporation of the graph, this approach leads to the ensuring of safety. 
This is a more explicit and natural way to achieve safety than in our previous work \cite{LemgoRL}, where a safety layer was used. We benchmark our approach against our previous work in a realistic simulation environment. 
The experimental results show that our approach leads to faster convergence and helps to avoid getting stuck in a local optimum during training. Also, the performance of the controller improves.

There is little research on how to ensure traffic-psychological pleasant control behavior in RL-based TSC. We tackle this problem by extending the action masking mechanism to include manually defined rules, that ensure such behavior. Experimental results show that this even further improves the training in terms of convergence and stability. However, due to the limitations of the RL controller imposed by the action masking mechanism, the performance is slightly worse. This indicates that a reasonable balance between limiting rules and controller performance must be found when using action masking. We believe that our novel approach is a critical step towards the real-world application of RL-based TSC. 




The remainder of this paper is structured as follows:
Section~\ref{sec:preliminaries} provides background information on RL and the simulation environment we use in this work. In Section~\ref{sec:main}, we explain our approach of using action masking for ensuring safety and psychologically pleasant behavior. This approach is evaluated in a realistic simulation environment in Section~\ref{sec:evaluation}. Section~\ref{sec:conclusion} concludes the paper.

\section{Preliminaries}
\label{sec:preliminaries}
\subsection{Background on Reinforcement Learning}
Reinforcement learning algorithms optimize the behavior of \textit{agents} that are acting within an \textit{environment}. In most cases, they are formalized as a \textit{Markov decision process} (MDP), which is a tuple composed of the following elements \cite{LemgoRL}:
\begin{enumerate}
    \item A set of \textit{states} $\mathcal{S}$, where a state $s_t$ encodes information about the current situation of an environment at a given discrete time step $t$.  
    \item A set of \textit{actions} $\mathcal{A}$. At each time step, the agent selects an action $a_t$ to execute in the environment.
    \item A \textit{state transition probability} function $T(s_t, a_t, s_{t+1})$. 
    Given the current state $s_t$ and action $a_t$, this function defines a probability distribution over the successor state $s_{t+1}$.  
    \item A \textit{reward} function $R(s_t,a_t)$.
    Based on the action $a_t$ selected in state $s_t$, the agent receives a scalar reward signal $r_{t+1}$, indicating how good or bad the immediate effect of the action was.  
    \item A \textit{discount factor} $\gamma \in [0, 1]$, that sets the discounting of future rewards.
\end{enumerate}
The behavior of the agent is determined by its \textit{policy} $\pi$, that maps states to actions $\pi(a_t | s_t)$. The goal of an RL agent is to behave in a way, that the expected cumulative reward 
is maximized in the long run.

\subsection{Simulation Environment}
\begin{figure}
    \centering
    \includegraphics[width=0.48\textwidth]{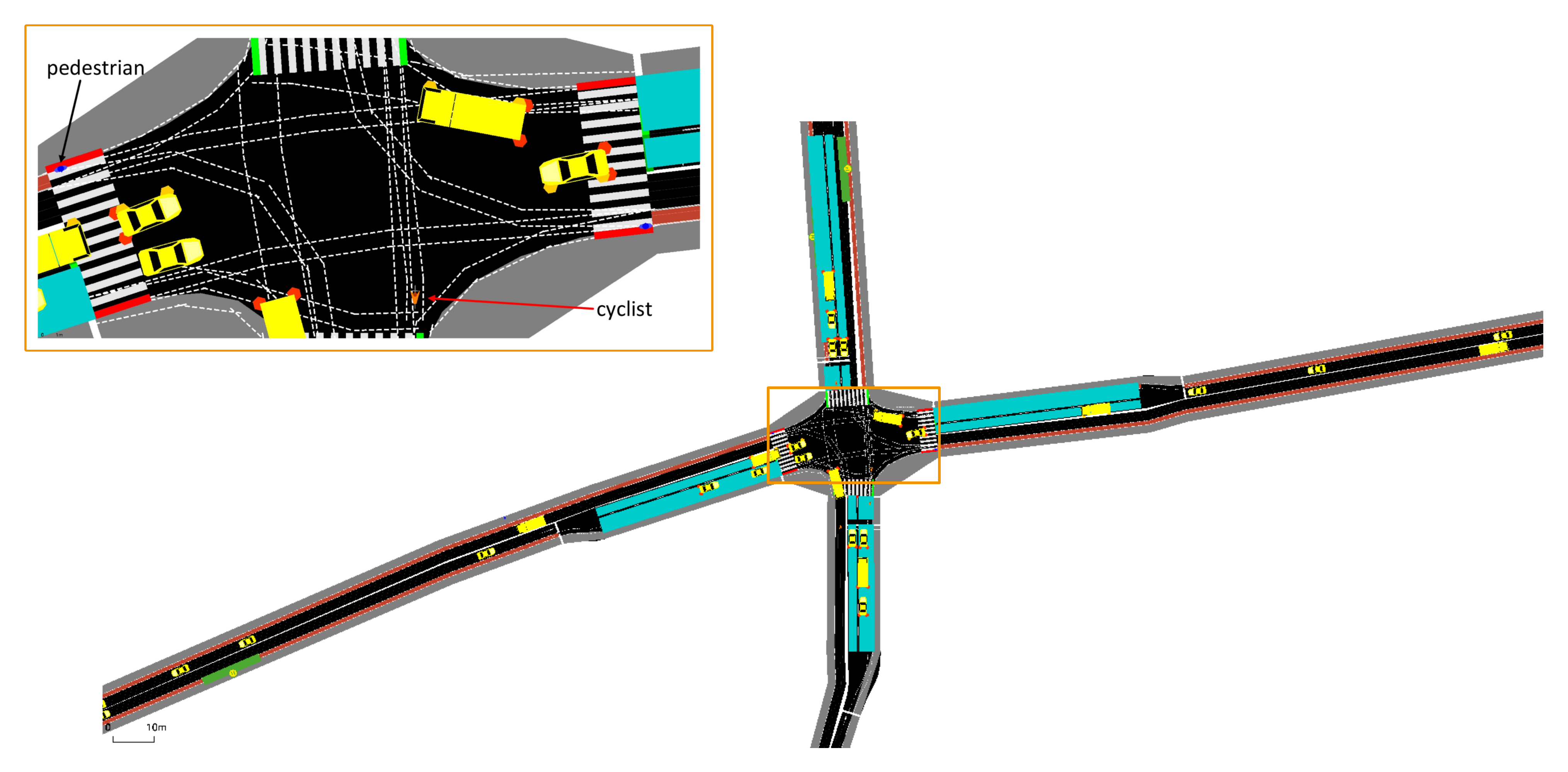}
    \caption{Simulation model of the OWL322 intersection}
    \label{fig:owl322_model}
\end{figure}

We aim to advance RL for TSC towards real-world applicability. Therefore, we need a realistic simulation environment in which we can integrate and demonstrate our solution. We use \textit{LemgoRL} \cite{LemgoRL} for this purpose. This is an open-source tool to train and evaluate RL agents as traffic signal controllers in a realistic simulation environment of Lemgo, a medium-sized town in Germany.
LemgoRL encompasses a simulation model of the \enquote{OWL322} intersection (see Fig.~\ref{fig:owl322_model}) built with the microscopic traffic simulation software SUMO \cite{Lopez2018}. 
The traffic data for the simulation is taken from a measurement of a typical afternoon rush hour.

\section{Safe and Psychologically Pleasant Traffic Signal Control with Reinforcement Learning}
\label{sec:main}

To achieve safe and psychologically pleasant control behavior of our RL policy, we focus on traffic phases. They will be explained first in Section~\ref{subsec:phases}. Next, we show how the information and relationships of all phases to each other can be modeled as a temporal, directed graph (Section~\ref{subsec:graph}). In Section~\ref{subsec:actionspace}, we motivate the design choices made for the action space.
Section~\ref{subsec:actionmasking} outlines how to use action masking to ensure safety and psychologically pleasant control behavior.
Lastly, we elaborate on the mathematical integration of action masking into an RL algorithm (Section~\ref{subsec:math}).

\subsection{Traffic Phases and Transitions}
\label{subsec:phases}
In most modern traffic signal controllers, \textit{traffic phases} are at the core of the control program. According to \cite{FGSV2015}, a traffic phase is defined as a certain compound of all traffic light signal colors that represent a basic state. 
In a traffic phase, non- or partially conflicting traffic flows are allowed to move, while the other road users have to wait. Table~\ref{tab:phases} shows the corresponding signal colors of all traffic phases for LemgoRL. Phase 3, for example, would allow vehicle traffic only coming from west and east, while vehicles coming from north and south, as well as pedestrians, have to wait.

\begin{table}
\renewcommand{\arraystretch}{1.3}
\centering
\begin{threeparttable}
\caption{Traffic light phases for the OWL322 intersection \cite{LemgoRL}}
\label{tab:phases}
 \begin{tabular}{|c || c | c | c | c| c | c | } 
 \hline
 \bfseries phase & \bfseries veh & \bfseries veh & \bfseries veh & \bfseries veh & \bfseries ped & \bfseries ped \\ 
 \bfseries nr & \bfseries west & \bfseries east & \bfseries north & \bfseries south & \bfseries w\&e & \bfseries n\&s \\ 
 \hline\hline
 1 & r & r & r & r & r & r \\ 
 \hline 
 2 & g & g & r & r & r & g \\ 
 \hline
 3 & g & g & r & r & r & r \\ 
 \hline
 4 & g & r & r & r & r & r \\ 
 \hline
 5 & g+left~g & r & r & r & r & r \\ 
 \hline
 6 & r & r & g & g & g & r \\ 
 \hline
 7 & r & r & g & g & r & r \\ 
 \hline
 8 & r & r & g & r & r & r \\ 
 \hline
\end{tabular}
\begin{tablenotes}
\item [*]vehicle traffic signal encompasses all directions (e.g. veh~west=all vehicles coming from the west and traveling to the north, east, or south) 
\item [**]r=red, g=green, left~g=left turning vehicles have priority green 
\end{tablenotes}
\end{threeparttable}
\end{table}

Whenever the controller requires a different phase to be active than the current one, a \textit{phase transition} is initiated. During a phase transition, the green signals first become yellow and then red for a certain amount of time. Afterward, the green signals of the new phase get active. Determining the required timing and order of green signals becoming yellow and then red during a phase transition is a traffic engineering task. It implies calculating the so-called intergreen times for \enquote{all combinations of conflicting traffic flows} \cite{FGSV2015}. The intergreen times \enquote{in turn depend on the permissible speed, the road user groups and actual topology and geometrical dimensions of the intersection} \cite{LemgoRL}. 
For instance, a phase transition from phase 2 (vehicles west and east \& pedestrians north and south) to phase 3 (only vehicles west and east) requires phase 2 to be active for at least $10s$, so that pedestrians can safely cross the street before they get a red signal.

\begin{figure}
    \centering
    \includegraphics[width=0.48\textwidth]{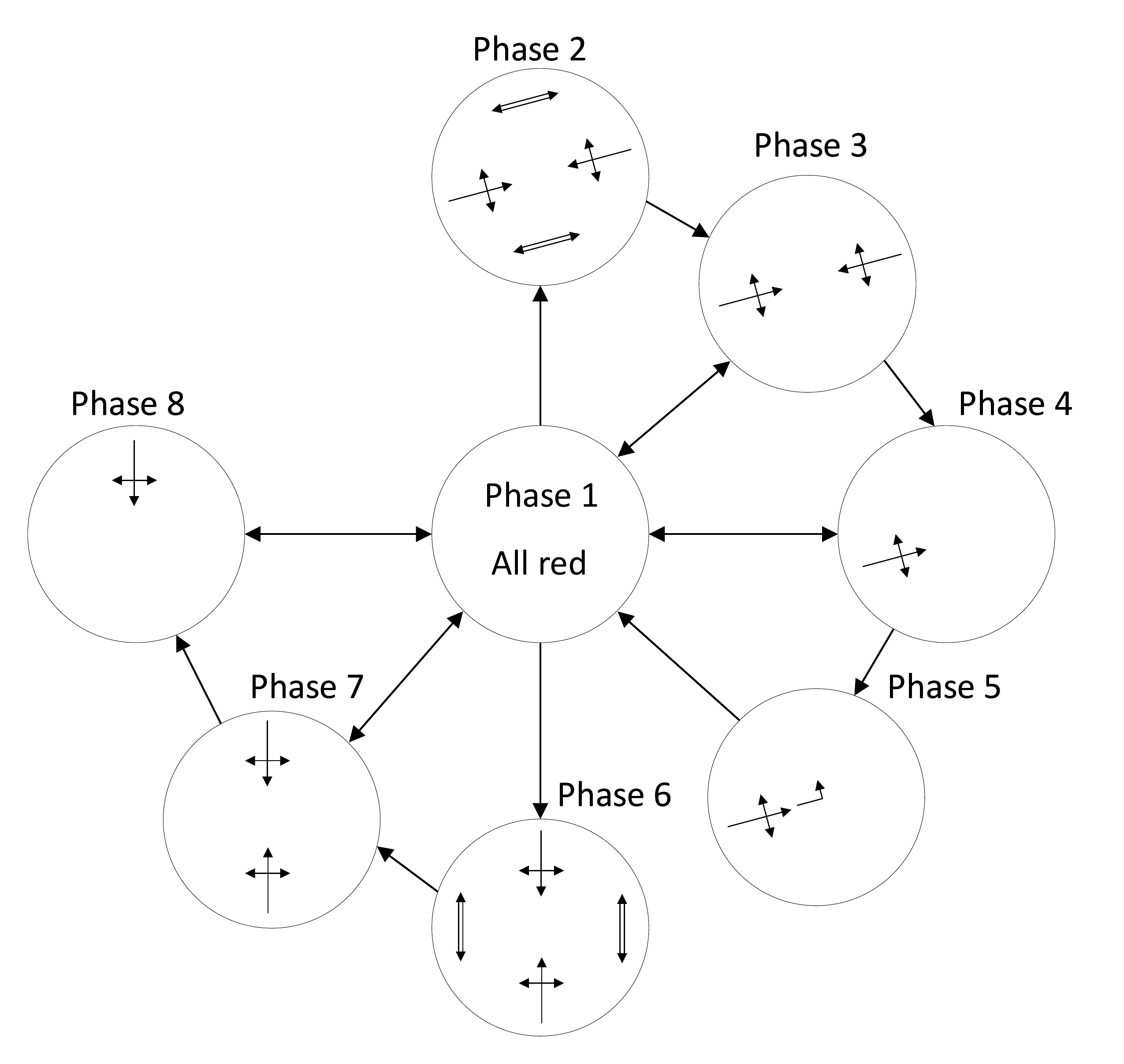}
    \caption{Traffic phases of LemgoRL and their transitions presented as a graph. The single-lined arrows represent the permitted directions of travel for vehicles and the double-lined arrows indicate that pedestrians are allowed to cross the street in the respective direction.}
    \label{fig:phase_diagram}
\end{figure}

Furthermore, in most cases, phases are only allowed to transit to a subset of all other phases. In Fig.~\ref{fig:phase_diagram} the traffic phases and all possible transitions of LemgoRL are presented as a directed graph, where the phases are nodes and the transitions are edges. For example, a phase transition from phase 3 (vehicles west and east) to phase 7 (vehicles north and south) requires phase 1 (all red) as an intermediate phase to be active for some seconds. That ensures, that all vehicles coming from west and east can safely leave the center of the intersection before north and south traffic is allowed to flow.   

\subsection{Temporal, Directed Graph and Safety Criteria}
\label{subsec:graph}

The information and relationships of all phases to each other and which and when phase transitions are possible can be modelled as a temporal, directed graph $G=(\mathcal{P},\mathcal{E},f)$ with discrete time steps t, where the phases and phase transitions constitute the node set $\mathcal{P}=\{p_1,p_2, ... p_N\}$ and the set of directed edges $\mathcal{E}=\{e_1,e_2,...,e_O\}$. For each edge $e_i = (p_j,p_k)$ where $p_j,p_k \in \mathcal{P}$, a safety function $f$ determines, whether the system can move along that edge or not: $$f(e_i, H(t), D(t)) = \begin{cases}
 1, \begin{aligned}[c]\text{      if the system can move} \\ \text{ along }  e_i \text{ at } t
\end{aligned}\\
0, \text{      otherwise,}
\end{cases}$$ 
where $H(t)=(h_0,h_{1},...,h_{t})$ represents the trajectory of phases up to the current phase $h_{t}$ and $D(t)=(d_0,d_{1},...,d_{t})$ their respective duration.

It's noteworthy, that we define the edge set to include also edges from each phase to itself. As long as the corresponding safety function has the value of $1$, it is allowed to stay in that phase. If the maximum time for this phase is exceeded, the safety function yields $0$ and the system has to move to another phase. 
Furthermore, we define a \textit{phase mask} $\bm{m}_G$
\begin{equation}
\label{eq:phase_mask_graph}
\bm{m}_G(t)=[m_{G,1}(t),m_{G,2}(t),...,m_{G,N}(t)]    
\end{equation}
where each element $m_{G,i}$ corresponds to a phase $p_i$. The phase mask provides information about which phase the system may access in the next time step. 
When there is at least one edge with $p_i$ as the target phase on which the system is allowed to move, then $m_{G,i}(t)=1$. This is true if $f((p_x,p_i), H(t), D(t))=1$ for $p_x \in \mathcal{P}$.
Otherwise, $m_{G,i}(t)=0$. This can be formulated as: 
\begin{equation}
\label{eq:phase_mask_graph_element}
m_{G,i}(t) = \bigvee_{p_x \in \mathcal{P}} f((p_x,p_i), H(t), D(t)) 
\end{equation}

The graph incorporates all the traffic engineering domain knowledge necessary to act safely. The goal of our approach is therefore to integrate this domain knowledge into the RL agent. As long as the RL agent operates within the constraints defined by $G$, the safety of the intersection traffic is guaranteed. 


\subsection{Action Space}
\label{subsec:actionspace}
The action space definition of the RL algorithm is key to integrating the traffic domain knowledge present in the graph. Action spaces can basically be divided into \textit{discrete} and \textit{continuous}.

An action $a$ of a \textit{discrete} action space is an integer with $a \in \{0,1,...,N\}$, 
where $N \in \mathbb{N}$. Each value represents a corresponding action the RL agent would choose in the environment. According to \cite{Haydari2020}, most approaches in RL for traffic signal control use discrete action spaces. 
Typically, an action represents one possibility of all discrete traffic phases, as it is done e.g. in \cite{Pol2016} or \cite{Gao2017}. A different option is to determine a fixed sequence of phases in advance and use a binary action of either staying in the current phase or switching to the next one (see e.g. \cite{Wei2018}).



In a \textit{continuous} action space, an action $a \in \mathbb{R}$ corresponds to a real number the agent would apply in the environment. In the context of RL-based TSC, only few applications with continuous action spaces are found \cite{Haydari2020}. 
For example, in \cite{Genders}, the action represents the duration of the next traffic phase, where the sequence of phases is predefined. Furthermore, the phase duration is limited between a specified minimum and maximum duration and is also rounded to the nearest integer. 


In our approach, we define a discrete action space that encompasses all traffic phases. Since in a traffic phase only non- or partially conflicting flows are allowed, this action space is safe-by-design. 
In the case of the LemgoRL environment, this leads to an action space of $a \in \{0,1,...,7\}$, where $a=0$ represents a transition to or staying in phase 1 and $a=1$ the same for phase 2 etc. The reason for not using a predefined sequence as it was done in the other two approaches is due to the loss of flexibility. For example in a fixed sequence, phase 3 would always precede phase 4. But in case of a much higher traffic demand coming from the west than from the east, skipping phase 3 and directly going to phase 4 reduces the volume of traffic more effectively.

\subsection{Action Masking}
\label{subsec:actionmasking}
To further leverage domain knowledge available in the graph, we use a technique from the field of \textit{action space shaping}. An overview about common action space shaping techniques can be found in \cite{Kanervisto2020}. In general, action space shaping transforms the action space of the environment with the goal to \enquote{make learning easier} \cite{Kanervisto2020}. For our purposes, we use a technique called \textit{action masking} \cite{Krasowski2020}, \cite{Huang2020}, which is referred to as \enquote{remove action} in \cite{Kanervisto2020}. In action masking, the action space is transformed through a mask $\bm{m}_{a}(t) = [m_{a,1}(t),m_{a,2}(t),...,m_{a,N}(t)]$, which is a vector with the elements defined as \begin{equation}
    m_{a,i}(t) = \begin{cases}
    0, & \text{if } a_i \text{ not available or desired} \\
    1, & \text{otherwise}
    \end{cases}
\end{equation} 
By applying the mask to the action space, the probabilities of unavailable or undesired actions are set to zero. Action masking makes the training more sample efficient since there are fewer actions for the agent to choose from, but it can also lead to worse performance \cite{Kanervisto2020}.


\subsubsection{Action Masking to Ensure Safety}
The goal of our approach is to make sure that the RL agent moves within the constraints of the graph $G$ so that safety is guaranteed. One part of the solution is to define a discrete action space encompassing the traffic phases. The other part, making sure that only allowed phase transitions are conducted, will be achieved through action masking. By using $\bm{m}_G$ defined in (\ref{eq:phase_mask_graph}) as mask for the RL agent 
\begin{equation}
\label{eq:action_mask_eq_graph_mask}
    \bm{m}_a=\bm{m}_G, 
\end{equation}
we ensure an action selection process that is in line with the constraints of the graph. 

\begin{figure}
    \centering
    \includegraphics[width=0.48\textwidth]{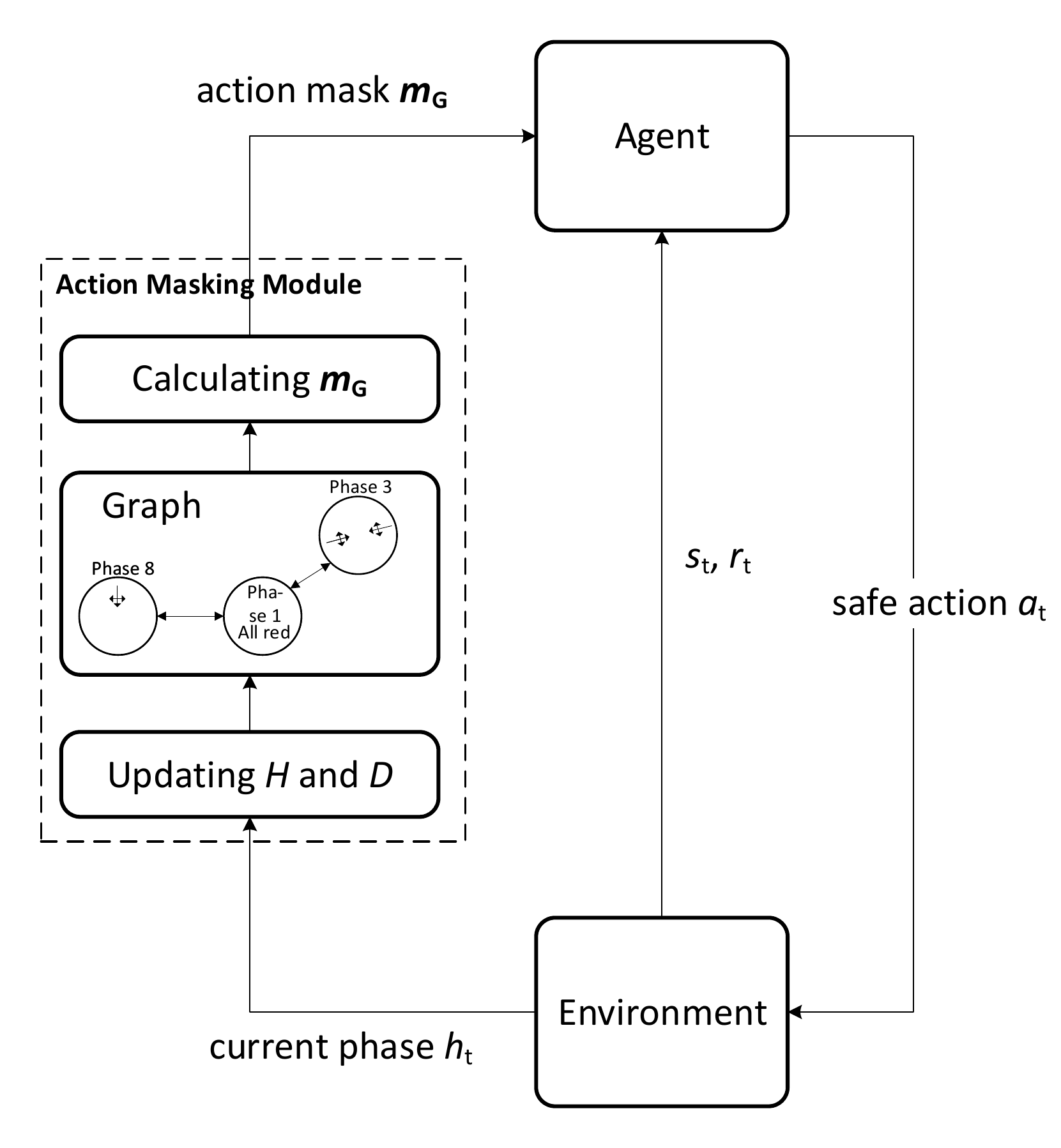}
    \caption{Reinforcement learning workflow with action masking}
    \label{fig:rl_action_masking}
\end{figure}

Fig.~\ref{fig:rl_action_masking} shows how action masking along with the graph is integrated into the RL workflow. At each time step, the environment provides $s_t$ and $r_t$ to the agent and additionally the current phase $h_t$ to the action masking module. Based on $h_t$, the phase trajectory $H$ and the corresponding phase duration $D$ is updated as well as the current state of the graph. Afterward, the safety function is calculated for every edge $e_i \in E$ to determine whether the system may move along that edge or not. Based on this information, the phase mask $\bm{m}_G$ respectively action mask $\bm{m}_a$ is determined. When choosing an action, the action mask sets all unsafe actions to probability zero, so that the agent can select safe phases only.

\begin{figure}
    \centering
    \includegraphics[width=0.48\textwidth]{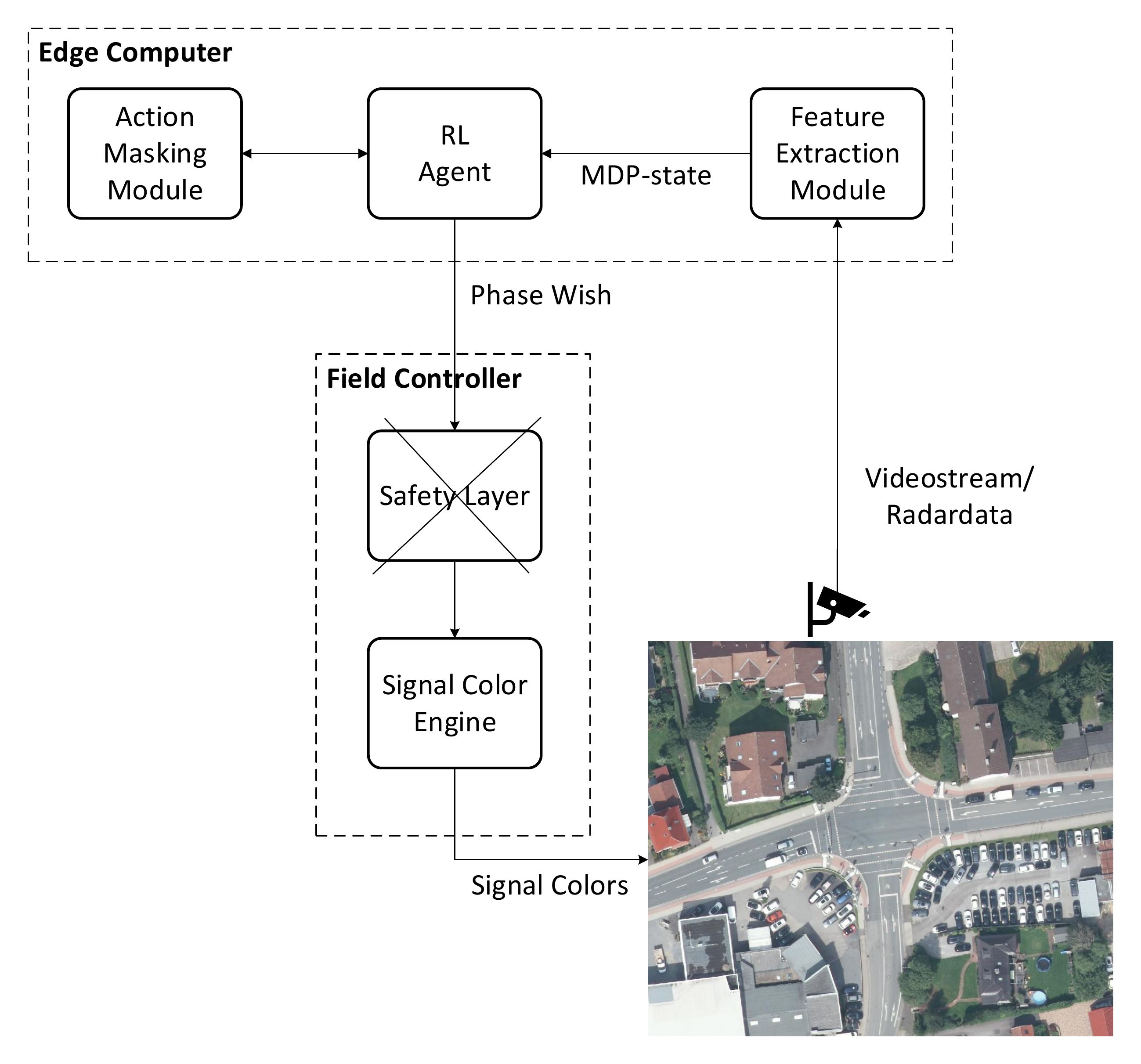}
    \caption{Real-world deployment architecture with action masking module instead of safety layer}
    \label{fig:real_world_deployment}
\end{figure}

By guaranteeing safety, the RL agent together with the action masking module could be deployed as real-world TSC. A possible architecture for real-world deployment is shown in Fig.~\ref{fig:real_world_deployment}. 
Data from camera or radar sensors are processed in real-time by a feature extraction module that transforms them into an MDP state. Based on that state, the RL agent selects the next phase. 
In our previous work \cite{LemgoRL}, a safety layer was needed to check compliance with all safety requirements. This is not necessary for our current approach, since the action masking module guarantees safe phase wishes. 
Therefore, the phase wish can be directly sent to the signal color engine, where it is converted into the signal colors for the traffic light devices at the intersection.

\subsubsection{Action Masking to Ensure Psychologically Pleasant Behavior}
Safety is one concern when it comes to the real-world deployment of RL-based TSC. Another one is the psychological impact on the road participants, more specifically the stress in drivers caused by the TSC. There are numerous works in literature dealing with the question of what causes stress in drivers and how this stress impacts the traffic situation \cite{Ishihara2001, Prasolenko2015, Westerman2000}. In general, it can be said, that driver stress \enquote{is highly correlated with the driving safety} \cite{Wang2019}. For that reason alone, an RL-based TSC should try to avoid stressful traffic situations.

We faced such situations during the training of RL agents for the LemgoRL environment. For example, when the system goes from phase 3 to phase 1, because the traffic coming from the west and east has drained away, and now the system wants to go to phase 6 or 7 in order to clear the traffic coming from the north and south. For safety reasons, the system first has to stay in phase 1 for more than 6 seconds. If in that time, the traffic demand from west and east grows very fast, the system goes back to phase 3 again instead of phase 6 or 7. I.e. a trajectory of $3\rightarrow 1 \rightarrow 3$ is occurring. This behavior might be optimal in order to achieve the highest cumulative reward, but in terms of the psychological effect on drivers, it is detrimental and should be avoided.

We introduce a natural solution to overcome such odd behavior by forbidding it via action masking. Therefore, we define a \textit{psychological action mask} $\bm{m}_{PS}(t) = [m_{PS,1}(t),m_{PS,2}(t),...,m_{PS,N}(t)]$ that gets extracted from a set of manually, predefined rules at every time step. These rules reflect traffic engineering domain knowledge to avoid stressful situations for the drivers.  
Then, we redefine (\ref{eq:action_mask_eq_graph_mask}) to include this set of rules via logical conjunction of $\bm{m}_G(t)$ and $\bm{m}_{PS}(t)$: 
\begin{align*}
    \bm{m}_a(t) & = \bm{m}_G(t) \land \bm{m}_{PS}(t) \nonumber\\
    & = (m_{G,1}(t) \land m_{PS,1}(t) ,..., m_{G,N}(t) \land m_{PS,N}(t)) 
    \label{eq:action_mask_bool}
\end{align*}
Now the action mask for the RL agent considers the safety rules extracted from the graph as well as the rules to avoid odd control behavior. 

\subsection{Algorithm Selection and Objective Function}
\label{subsec:math}
In order to determine the reinforcement learning algorithm we are going to integrate the action masking mechanism into, we compared the performance of three state-of-the-art RL algorithms on the LemgoRL environment without action masking. We included IMPALA \cite{IMPALA2018}, Ape-X DQN \cite{APEX2018}, and PPO \cite{Schulman2017} in our benchmark. As in our previous work \cite{LemgoRL}, we use the Ray RLlib \cite{Liang2018} implementation of these algorithms, since training can be done on several cores in parallel which reduces training time. It turned out, that PPO performs best. 

In the following, we are presenting the mathematical core elements of the PPO algorithm and how to integrate the action masking mechanism into it. 
Ray RLlib`s implementation of the PPO algorithm follows the original paper regarding the PPO objective $L_t(\theta)$ \cite{Schulman2017}:
\begin{equation}
    \label{eq:objective}
    L_t(\theta) =  \mathbb{\hat{E}}_t [L_t^{CLIP}(\theta)-c_1L_t^{VF}(\theta)+c_2S_{\pi_\theta}(s_t)], 
\end{equation}
with the clipped surrogate objective $L_t^{CLIP}$
\begin{equation*}
    L_t^{CLIP}(\theta)=\mathbb{\hat{E}}_t[\mathrm{min}(r_t(\theta)\hat{A}_t,\mathrm{clip}(r_t(\theta),1-\epsilon,1+\epsilon)\hat{A}_t)],
\end{equation*}
the squared-error loss $L_t^{VF}(\theta)=(V_\theta(s_t)-V_t^{targ})^2$
and the entropy bonus $S$\footnote{The explicit definition of the entropy bonus is not part of the original PPO paper, but the definition shown here corresponds to the implementation in Ray RLlib.}
\begin{equation}
\label{eq:entropy}
    S_{\pi_\theta}(s_t)=-\sum_{a_i \in \mathcal{A}} \pi_\theta(a_i | s_t)\mathrm{log}(\pi_\theta(a_i | s_t)),
\end{equation}
where $c_1$ and $c_2$ are coefficients. $\theta$ denotes the trainable parameters of the neural network. The core element of the PPO objective is $L_t^{CLIP}$ depending on the probability ratio  $r_t(\theta)$ 
\begin{equation}
\label{eq:prob_r}
    r_t(\theta)=\frac{\pi_\theta(a_t | s_t)}{\pi_{\theta_{\text{old}}}(a_t | s_t)},
\end{equation}
with $\pi_{\theta_{\text{old}}}$ being the old policy before the update, and the advantage function $\hat{A}_t$ calculated by a generalized advantage estimator \cite{GAE2018}. Clipping the probability ratio to $[1-\epsilon,1+\epsilon]$ where $\epsilon$ is a constant, prevents $\pi_{\theta}$ to move too far away from $\pi_{\theta_{\text{old}}}$ in the case the objective would improve.    

Since we choose a network architecture where the policy and value functions share parameters, we have to include $L_t^{VF}$ as value function error term in order to learn the state-value function $V_\theta(s)$. 
The target value function $V_t^{targ}$ needed for the error term is gathered through simulations. 
The purpose of including $S_{\pi_\theta}$ in the loss function is to stimulate more exploration and to prevent the agent from converging to suboptimal policies \cite{Mnih2013}.

Through the introduction of the mask $\bm{m}_a(t)$ in the action selection process, there is a new probability ratio $r_t^m$ based on (\ref{eq:prob_r}) 
\begin{equation*}
r_t^m(\theta)=\frac{\pi_\theta^m(a_t, \bm{m}_a(t) | s_t)}{\pi_{\theta_{\text{old}}}^m(a_t, \bm{m}_a(t) | s_t)}
\end{equation*}
with the modified policy values $\pi_{\theta}^m$ and $\pi_{\theta_{\text{old}}}^m$. 
Since the masking always ensures that $a_t$ is a safe and psychological pleasant action, 
the modified policy values equals the original ones $\pi_{\theta}$ and $\pi_{\theta_{\text{old}}}$. 
Therefore, $r_t^m(\theta)$ equals $r_t(\theta)$ \cite{Krasowski2020}. $\hat{A}_t$ is not modified by action masking either. As a consequence, $L_t^{CLIP}$ is not affected by action masking. The same is true for $L_t^{VF}$ since it only depends on the value function. For the entropy bonus, the situation is different. Based on (\ref{eq:entropy}), the entropy bonus becomes
\begin{align*}
\label{eq:entropy_m}
    S^m_{\pi_\theta}(s_t) & =-\sum_{a_i \in \mathcal{A}} \pi_\theta^m(a_i, \bm{m}_a(t) | s_t)\mathrm{log}(\pi_\theta^m(a_i, \bm{m}_a(t) | s_t)) \nonumber\\
    & =-\sum_{a_i \in \mathcal{A}_{\text{masked}}} \pi_\theta(a_i | s_t)\mathrm{log}(\pi_\theta(a_i | s_t))
\end{align*}
where $\mathcal{A}_{\text{masked}}$ is the set of possible actions at time t based on $\bm{m}_a(t)$. Since masking out an action $a_i$ by $m_{a,i}=0$ leads to $ \pi_\theta^m(a_i, \bm{m}_a(t) | s_t)=0$, $S^m_{\pi_\theta}$ corresponds to $S_{\pi_\theta}$ except that it is the entropy over $\mathcal{A}_{\text{masked}}$ and not $\mathcal{A}$. Thus, based on (\ref{eq:objective}) the new PPO objective becomes 
\begin{equation*}
    L_t^m(\theta) =  \mathbb{\hat{E}}_t [L_t^{CLIP}(\theta)-c_1L_t^{VF}(\theta)+c_2S_{\pi_\theta}^m(s_t)]. 
\end{equation*}
The gradient of the new objective is still a valid policy gradient as shown in \cite{Huang2020}.






\section{Evaluation}
\label{sec:evaluation}
\subsection{Experimental Setup}
In order to evaluate and quantify the impact of action masking on RL-based TSC, we conduct an experiment with 3 different controllers. 

\subsubsection{Controllers}
The controller (a) incorporates a safety layer that checks compliance of the agent's phase wishes with all safety requirements. A phase wish that is illegal leads to a no-op action, i.e. staying in that phase if the maximum allowed phase duration is not exceeded or going to the next phase otherwise. There is no action masking involved here. This controller is the baseline. In controller (b), we use action masking to mask out illegal, i.e. not safe actions. The agent, therefore, selects only safe actions, so that there is no need for a subsequent safety layer anymore. In controller (c), we extend the action masking mechanism to incorporate rules, that forbid psychologically inappropriate behavior. So in total, we use the following 3 controllers for the experiment:  
\begin{enumerate}
    \item[(a)] PPO \textit{with} safety layer and \textit{without} action masking (baseline)
    \item[(b)] PPO \textit{without} safety layer and \textit{with} action masking for safety only
    \item[(c)] PPO \textit{without} safety layer and \textit{with} action masking for safety and psychologically pleasant behavior
\end{enumerate}


\subsubsection{Evaluation Methodology and Metrics}
To ensure a fair comparison, we train each RL agent for each controller 5 times with different random seeds while we use the same set of random seeds for each controller. This makes in total 15 agents. The training is done in the LemgoRL framework. 

We evaluate the controllers in 2 ways. 
First, we compare the \textit{learning curves}
by considering the \enquote{the speed of learning (how fast curve rises), ... the final performance (how high curve gets) and if the agent learns at all (if the curve rises)} \cite{Kanervisto2020}. 
Second, we perform 10 simulation runs with each of the 15 agents. Each run has a different random seed to bring in variations in the traffic demand, but for each agent, the same set of random seeds is used to ensure a fair comparison. We calculate the average over all lanes respectively pedestrian crosswalks and all 10 runs for the following traffic metrics: 
\begin{enumerate}
    \item \texttt{queue}: The length of the queue of waiting vehicles (velocity $< 0.1\text{m/s}$) in a lane.
    \item \texttt{wait\_veh}: The maximum time a vehicle waits in a lane when the signal color is red. 
    \item \texttt{speed}: The average speed of all vehicles in a lane.
    \item \texttt{wait\_ped}: The maximum time a pedestrian waits for a green signal at a crosswalk.
    \item \texttt{stops}: The avg. number of stops of a vehicle while it approaches the intersection.
    \item \texttt{travel\_time}: The avg. time for a vehicle to cross the intersection.
\end{enumerate}
Also, we compute the \texttt{cumulative reward} a controller collects during a run.

\subsubsection{Reinforcement Learning}
For all RL agents we use the PPO algorithm with the hyperparameters specified in Table~\ref{tab:ppo_hp}. The state definition is inspired by \cite{LemgoRL}. It is defined as a vector containing the vehicle-related information \texttt{queue}, \texttt{wait\_veh}, \texttt{speed}, and \texttt{wave} for each lane of the intersection, where \texttt{wave} is defined as the number of vehicles in a lane. Additionally, the state vector includes \texttt{wait\_ped} for each crosswalk, the one-hot encoded current phase, and the elapsed time in the current phase. This vector is fed into a fully connected neural network with 2 layers, each layer having 128 neurons. For the reward function, we extend the definition in \cite{LemgoRL} to include $\texttt{stops\_total}_t$, the number of all vehicles stops in a lane at a given time $t$:    
\begin{multline*}
    r_{t+1} = - \sum_{l} \Bigl(\alpha_{q}\cdot \texttt{queue}_{t+1}[l]  +  \alpha_{w,veh}\cdot \texttt{wait\_veh}_{t+1}[l] \\ + \alpha_{w,ped}\cdot \texttt{wait\_ped}_{t+1}[l] + \alpha_{st}\cdot \texttt{stops\_total}_{t+1}[l]\Bigr),
\end{multline*} 
where $l$ denotes an incoming lane or pedestrian crosswalk of the intersection and $\alpha_{q}$, $\alpha_{w,veh}$, $\alpha_{w,ped}$, and $\alpha_{w,st}$ are coefficients for weighting the individual components. The values of the coefficients are shown in Table~\ref{tab:ppo_hp}. Since the reward is defined as the negative sum of $\texttt{queue}$, $\texttt{wait\_veh}$, $\texttt{wait\_ped}$, and $\texttt{stops\_total}$ over all lanes respectively pedestrian crosswalks, the agent tries to reduce this metrics, which leads to fluent traffic flow.

\begin{table}[ht]
\centering
\caption{Hyperparameters used for PPO algorithm for all three controllers (nomenclature follows RLlib)}
\label{tab:ppo_hp}     
\begin{tabular}{l l}
\hline
\textbf{Hyperparameter} & \textbf{Value} \\
\hline
train\_batch\_size & $2048$ \\
num\_sgd\_iter & $20$ \\
gamma & $0.98$ \\
lambda & $0.95$ \\
vf\_loss\_coeff & $0.005$ \\
lr & $5e-5$ \\
activation function & tanh \\
fcnet\_hiddens* & $[128, 128]$ \\
optimizer & Adam \\
episode\_total & $2500$ \\
\hline 
$\alpha_{q}$ & $0.8$\\
$\alpha_{w,veh}$ & $0.8$\\
$\alpha_{w,ped}$ & $0.2$\\
$\alpha_{w,st}$ & $1.9$\\
\hline
\end{tabular}
\begin{tablenotes}
\centering
\item [*]*size of the neural network
\end{tablenotes}
\end{table}


\subsection{Results and Discussion}
\begin{figure*}[!ht]
    \centering
    \begin{subfigure}{0.31\textwidth} 
        \centering
        \includegraphics[width=0.99\textwidth]{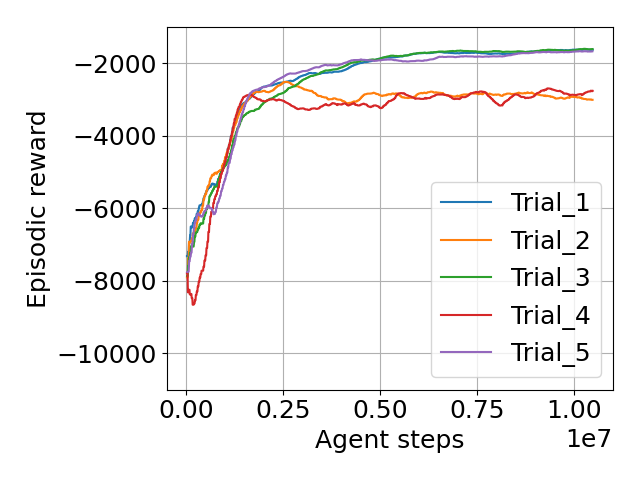}
        \caption{}
        \label{fig:exp80}
    \end{subfigure}%
    \begin{subfigure}{0.31\textwidth} 
        \centering
        \includegraphics[width=0.99\textwidth]{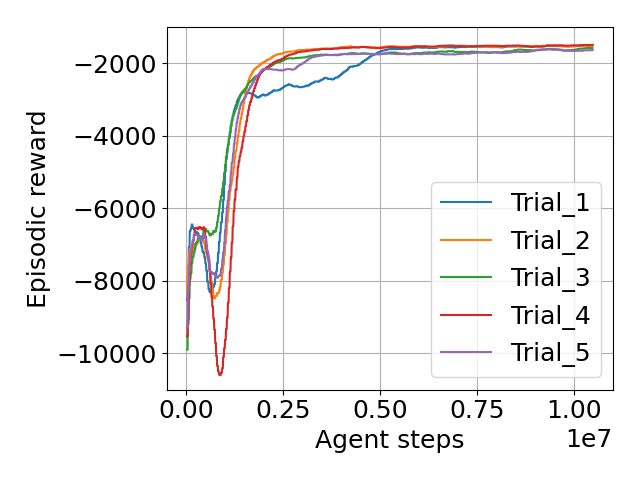}
        \caption{}
        \label{fig:exp81}
    \end{subfigure}%
    \begin{subfigure}{0.31\textwidth} 
        \centering
        \includegraphics[width=0.99\textwidth]{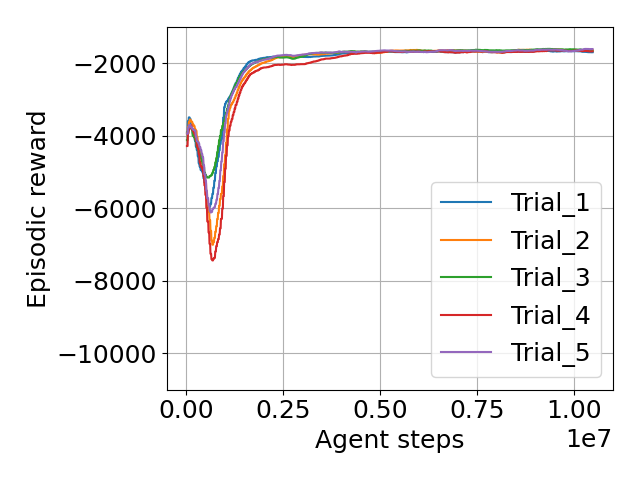}
        \caption{}
        \label{fig:exp82}
    \end{subfigure}
    
    \caption{Learning curves for all controllers}
    \label{fig:exp80-82}
\end{figure*}

The learning curves for all controllers and all trials are shown in Fig.~\ref{fig:exp80-82} and the results from the simulation runs are shown in Table~\ref{tab:metrics}.

\begin{table*}[!ht]
    \centering
      \caption{Traffic metrics and cumulative reward. The traffic metrics $\texttt{queue}$, $\texttt{wait\_veh}$, $\texttt{speed}$, and $\texttt{wait\_ped}$ are averaged over all lanes and pedestrian crosswalks of the intersection. All metrics for a trial are averaged over 10 simulation runs. The best values of a metric for a controller are highlighted in green and the second-best in yellow.}
    \label{tab:metrics} 
    \begin{tabular}{c|c|c|c|c|c|c|c|c}
    \hline
        ctrl. & trial & \makecell[c]{$\texttt{queue [m]}$} & \makecell[c]{$\texttt{wait\_veh [s]}$} & \texttt{speed [m/s]} & \makecell[c]{$\texttt{stops}$} & \makecell[c]{$\texttt{travel\_time [s]}$} & \makecell[c]{$\texttt{wait\_ped [s]}$} & \makecell[c]{\texttt{cum.} \\ \texttt{reward}} \\ \hline
        \multirow{6}{*}{(a)} & 1 & \cellcolor{yellow!30}54.8 & \cellcolor{yellow!30}62.7 & \cellcolor{green!30}9.5 & \cellcolor{green!30}0.74 & \cellcolor{yellow!30}49.9 & 14.6 & \cellcolor{yellow!30}-750.6 \\ 
         & 2 & 76.1 & 74.1 & 7.4 & 1.09 & 60.3 & \cellcolor{green!30}13.5 & -1331.7 \\ 
         & 3 & \cellcolor{green!30}53.2 & \cellcolor{green!30}59.5 & \cellcolor{green!30}9.5 & \cellcolor{yellow!30}0.76 & 56.6 & \cellcolor{yellow!30}13.6 & \cellcolor{green!30}-739.8 \\ 
         & 4 & 74.7 & 70.2 & 7.7 & 1.11 & \cellcolor{green!30}49.7 & 13.7 & -1233.8 \\ 
         & 5 & 57.0 & 64.4 & \cellcolor{yellow!30}9.4 & 0.77 & 52.8 & 14.2 & -756.8 \\ 
         \cline{2-9}
         & \textbf{avg. best 3} & \textbf{55.0} &	\textbf{62.2} &	\textbf{9.5} &	\textbf{0.76} &	\textbf{53.1} &	\textbf{14.1} &	\textbf{-749.1} \\ 
         & std &	10.09 &	5.26 &	0.97 &	0.17 &	4.09 &	0.39 &	263.31 \\ \hline
                    
        \multirow{6}{*}{(b)} & 1 & \cellcolor{yellow!30}51.2 & \cellcolor{green!30}51.1 & \cellcolor{green!30}9.8 & \cellcolor{yellow!30}0.76 & \cellcolor{green!30}49.2 & 19.4 & \cellcolor{green!30}-685.7 \\ 
         & 2 & \cellcolor{green!30}50.3 & \cellcolor{yellow!30}51.7 & \cellcolor{green!30}9.8 & 0.77 & 51.7 & 21.5 & \cellcolor{yellow!30}-713.8 \\ 
         & 3 & 52.3 & 56.9 & 9.6 & \cellcolor{green!30}0.74 & \cellcolor{yellow!30}50.5 & \cellcolor{yellow!30}15.1 & -721.8 \\ 
         & 4 & 51.5 & 52.3 & \cellcolor{yellow!30}9.7 & 0.78 & 51.0 & 19.9 & -724.9 \\ 
         & 5 & 53.4 & 59.2 & 9.5 & \cellcolor{green!30}0.74 & 57.4 & \cellcolor{green!30}14.8 & -757.5 \\
         \cline{2-9}
         & \textbf{avg. best 3} &	\textbf{51.3} &	\textbf{53.3} &	\textbf{9.7} &	\textbf{0.76} &	\textbf{50.5} &	\textbf{18.7} &	\textbf{-707.1} \\ 
        & std &	1.05 &	3.21 &	0.10 &	0.02 &	2.84 &	2.69 &	22.98 \\ \hline
    
        \multirow{6}{*}{(c)} & 1 & 55.8 & 63.5 & 9.3 & 0.77 & 54.3 & \cellcolor{green!30}12.2 & -764.8 \\ 
         & 2 & \cellcolor{yellow!30}53.6 & \cellcolor{yellow!30}57.5 & \cellcolor{yellow!30}9.5 & 0.77 & \cellcolor{green!30}50.8 & \cellcolor{yellow!30}14.2 & \cellcolor{yellow!30}-747.7 \\ 
         & 3 & \cellcolor{yellow!30}53.6 & 61.6 & \cellcolor{green!30}9.6 & \cellcolor{green!30}0.72 & 51.9 & 15.8 & -753.9 \\ 
         & 4 & 53.8 & 59.2 & \cellcolor{yellow!30}9.5 & \cellcolor{yellow!30}0.75 & 55.4 & 14.3 & -770.7 \\ 
         & 5 & \cellcolor{green!30}53.2 & \cellcolor{green!30}56.1 & \cellcolor{yellow!30}9.5 & 0.77 & \cellcolor{yellow!30}51.5 & 14.4 & \cellcolor{green!30}-747.2 \\
         \cline{2-9}
         & \textbf{avg. best 3} &	\textbf{53.5} &	\textbf{58.4} &	\textbf{9.5} &	\textbf{0.75} &	\textbf{51.4} &	\textbf{14.8} &	\textbf{-749.6} \\ 
        & std &	0.93 &	2.67 &	0.08 &	0.02 &	1.75 &	1.15 &	9.41 \\ \hline
    \end{tabular}
\end{table*}

First, we investigate the difference between (a) and (b), i.e. the difference between using a safety layer or instead action masking to ensure safety. The most significant difference is that 2 of 5 agents from (a) get stuck in a local optimum around an episodic reward of $-3000$, while the other 3 agents from (a) and all 5 agents from (b) reach a level of around $-1800$.
Another aspect is the faster convergence when using action masking. In (b), 3 of 5 agents reach a performance over $-2000$ in less than approx. 2.4 million steps, but in (a) 3 of 5 agents need around 4.3 million steps to reach the same level. 
Furthermore, agents trained with action masking reach a slightly higher cumulative reward. This can be seen in the average cumulative reward for the best three agents of the controllers in Table~\ref{tab:metrics}. While in (a) this value is $-749.1$, in (b) the cumulative reward is $-707.1$.  
The data demonstrate that the usage of action masking results in avoiding local optima, faster convergence, and better performances of the agents. This can be explained through the reduced number of possible actions, that come with action masking, and that leads to more efficient exploration.
Considering the traffic metrics in Table~\ref{tab:metrics} gives a mixed picture. On the one hand, the controller using action masking outperforms the safety layer version in all vehicle-related metrics (for $\texttt{stops}$ the values are equal). But on the other hand, the safety layer version emphasizes more on reducing the waiting time for pedestrians, so that the average of this metric for the best three controllers in (a) amounts to $14.1 s$, while it is $18.7 s$ for (b). Thus, the agents in (b) have learned to prioritize vehicles at the expense of pedestrians which leads to a higher cumulative reward. 

In the following, we investigate the impact of extending the action masking mechanism to incorporate non-safety-related rules, so we compare the controllers of (b) and (c). One finding is, that the learning curves in (c) are more alike than those of (b). I.e. the shape of the curves, as well as the values, differ less from each other. 
An interesting finding was that there is a drop in the episodic reward at the beginning of the training in (b) and (c), but not in (a). The agents in (b) have shown a worse drop (trial 4 up to approx. $-10500$) than those of (c) (trial 4 up to approx. $-7500$). Why there is a drop when integrating action masking remains a subject for further research.
Furthermore, the convergence of the agents in (c) was slightly faster than those in (b). 
These findings are consistent with the interpretation of our previous analysis of comparing (a) and (b). Integrating more rules in the action masking mechanism leads to a more efficient exploration (reduced number of actions) and therefore to faster convergence and more stable training.
In addition, the standard deviations of all traffic metrics in Table~\ref{tab:metrics} are lower in (c) than in (b) or equal (in case of $\texttt{stops}$).
The performance of the control behavior is therefore less dependent on the initial training conditions when using action masking.  
However, there is one downside. By integrating more rules, the agent has less flexibility. In our case, this leads to slightly worse performance of the agents in (c). While the average cumulative reward of the best three controllers from (b) is $-707.1$, this value is $-749.6$ for (c).   

\section{Conclusion}
\label{sec:conclusion}
In this paper, we show an easy and practical approach to develop an RL-based traffic signal controller that acts at all times safe and can therefore be deployed in a real-world setting. We define a discrete action space that encompasses all possible traffic phases, which represent combinations of signal colors that allow non- or partial conflicting traffic flows. Most importantly, we extract traffic engineering domain knowledge about the interaction between phases as a temporal, directed graph and incorporate this graph via action masking into the RL algorithm. The action space is therefore safe-by-design. Empirical data gathered in an experiment with a realistic simulation scenario demonstrates the superiority of this approach over using a safety layer. Agents trained with our approach converge faster, reach a higher end performance, and avoid getting stuck in a local optimum.      

In the next step, we extend the action masking mechanism to incorporate rules, which are not directly related to safety, but to the psychological impact on drivers. Without any restriction, the RL agent could produce phase trajectories that might be optimal in terms of optimizing the cumulative reward, but that has a negative impact on the psychological state of traffic participants. Therefore, they should be avoided. Our experiment has shown that incorporating such rules lead to even faster convergence and more stable training. But as a drawback, the performance of that agents is slightly lower, since the agents have less flexibility in the action selection process. For this reason, care must be taken to find a sensible balance between restricting the agent and performance. 

By ensuring safety and psychologically pleasant control behavior, we believe that our approach is a major step towards the deployment of RL-based traffic signal controllers to real-world scenarios. In our future work, we therefore plan to deploy this solution at the OWL322 intersection in Lemgo.  



\bibliographystyle{IEEEtran.bst} 
\balance
\bibliography{paper2.bib}

\end{document}